\documentclass[11pt,a4paper]{article}
\usepackage[hyperref]{acl2019}
\usepackage{times}
\usepackage{latexsym}
\usepackage{lastpage,fancyhdr,graphicx}
\usepackage{epstopdf}

\usepackage{amsmath}
\usepackage{algorithm}
\usepackage[noend]{algpseudocode}
\usepackage{array}

\usepackage{url}
\usepackage{breqn}

\aclfinalcopy 

\title{Deep Two-path Semi-supervised Learning for Fake News Detection}

\author{Xishuang Dong, ~ Uboho Victor, ~  Shanta Chowdhury, ~Lijun Qian\\
 CREDIT, \\
 Department of Electrical and Computer Engineering, \\
 Prairie View A\&M University, Texas A\&M University System, Prairie View, Texas 77446 \\
 \texttt{\{xidong, liqian\}@pvamu.edu, uboho.dpc@outlook.com, shanta.chy10@gmail.com} }

\date{}

\begin{document}
\maketitle
\begin{abstract}
News in social media such as Twitter has been generated in high volume and speed. However, very few of them can be labeled (as fake or true news) in a short time.
In order to achieve timely detection of fake news in social media,  a novel deep two-path semi-supervised learning model is proposed, where one path is for supervised learning and the other is for unsupervised learning. These two paths implemented with convolutional neural networks are  jointly optimized to enhance detection performance. In addition, we build a shared convolutional neural networks between these two paths to share the low level features. Experimental results using Twitter datasets show that the proposed model can recognize fake news  effectively with very few labeled data.
  
 \end{abstract}
 
 \begin{figure*} [ht]
	\includegraphics[width=\linewidth]{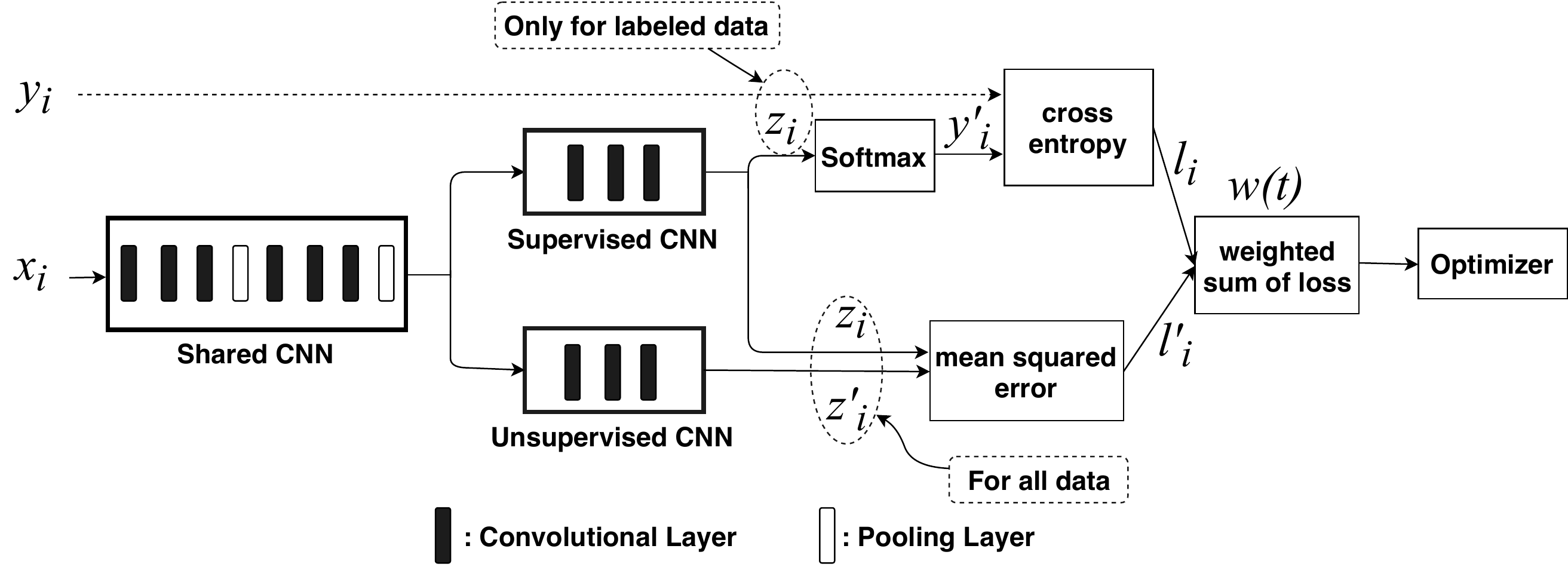}
	\caption{Framework of deep two-path semi-supervised learning (DTSL). Samples $x_{i}$ are inputs. Labels $y_{i}$ are available only for the labeled inputs and the associated cross-entropy loss component is evaluated only for those. $z_{i}$ and $z'_{i}$ are outputs from the supervised CNN and the unsupervised CNN, respectively. $y'_{i}$ is the predicted label for $x_{i}$. $l_{i}$ is the cross-entropy loss and $l'_{i}$ is the mean squared error loss. $w(t)$ are the weights for joint optimization of $l_{i}$ and $l'_{i}$. In the shared CNN, the first three convolutional layers each contains 128 ($3\times3$) filters, and the rest three convolutional layers each contains 256 ($3\times3$) filters. The three convolutional layers of the supervised CNN and those of the unsupervised CNN have the same configuration with  512 ($3\times3$) filters, 256 ($3\times3$) filters, and 128 ($3\times3$) filters. We use ($2\times2$) max-pooling for all pooling layers.}
	\label{Fig1_framework}
\end{figure*}

\section{Introduction}
Social media (e.g., Twitter, Facebook, and Weibo) has become a new ecosystem for spreading news \cite{pennycook2019fighting}. Nowadays, people are relying more on social media services rather than traditional media because of its advantages such as social awareness, global connectivity, and real-time sharing of digital information. 
Unfortunately, social media is full of fake news. 
Fake news consists of information that is intentionally and verifiably false to mislead readers, which is motivated by chasing personal or organizational profits~\cite{shu2017fake}. It has many similarities with spam messages since they share common features such as grammatical mistakes, false information, using similar limited set of words. Understanding what can be done to discourage fake news is of great importance.
For example, fake news has been propagated on Twitter like infectious virus during the 2016 election cycle in the United States \cite{allcott2017social, grinberg2019fake}. 

One of the fundamental steps to discourage fake news would be fake news detection.
Fake news detection \cite{hovy2016enemy, wang2017liar, potthast2018stylometric} is to determine the truthfulness of the news by analyzing the news contents and related information such as propagation patterns. Currently, as an emerging natural language processing (NLP) task, fake news detection has been solved with various models by analyzing the contents of fake news \cite{oshikawa2018survey}. Specifically, deep learning based fake news detection achieved good performance on different datasets, where both recurrent neural networks (RNN) and convolutional neural networks (CNN) are employed to construct supervised learning models to recognize fake news \cite{rashkin2017truth, long2017fake, wang2017liar}. However, 
since news spreads on social media at very high speed when an event happens, 
only very limited labeled data is available in practice for fake news detection, which is inadequate  for the supervised model to perform well.

In this paper, we propose a deep semi-supervised learning model by building two-path convolutional neural networks to accomplish fake news detection in the case of limited labeled data, where the model framework is shown in Figure \ref{Fig1_framework}. It consists of three components, namely, shared CNN, supervised CNN, and unsupervised CNN. One path is composed of shared CNN and supervised CNN while the other is made of shared CNN and unsupervised CNN. 
All data will go through both paths in the learning process, and generate the mean squared error loss, while only labeled data will be used to calculate the cross-entropy loss. Then a weighted sum of the two losses is used to optimize the proposed model.
We validate the proposed model on detecting fake news from tweets and experimental results demonstrate the effectiveness of the proposed method even with limited labeled tweets.

In summary, the contributions of this study are as follows:
\begin{itemize}

\item We proposed a novel deep two-path semi-supervised learning (DTSL) model containing three CNNs, where both labeled data and unlabeled data can be used jointly to train the model and enhance the detection performance. 

\item We validate our proposed model by testing on the PHEME dataset \cite{zubiaga2016analysing}  and observe that the proposed model perform better than supervised learning models when the training datasets and testing datasets don't share the same distribution since the proposed model will not be overfitting like the supervised learning models.



\end{itemize}

\begin{algorithm*}[ht]
	\caption{Learning in the proposed model}\label{euclid}
	\begin{algorithmic}[1]
		\Require{$x_{i}$ = training sample}
		\Require{$S$ = set of training samples}
		\Require{$y_{i}$ = label for labeled $x_{i}$ $i \in S$}
		\Require{$f_{\theta_{shared}}(x)$ = shared CNN with trainable parameters  $\theta_{shared}$}
		\Require{$f_{\theta_{sup}}(x)$ = supervised CNN with trainable parameters  $\theta_{sup}$}
		\Require{$f_{\theta_{unsup}}(x)$ = unsupervised CNN with trainable parameters  $\theta_{unsup}$}
		\Require{$w(t)$ = unsupervised weight ramp-up function}
		
		\For{$t$ in~[1, num epochs] }
 			 \For{each minibatch $B$}
      				\State{$z_{i \in B} \gets f_{\theta_{sup}}({f_{\theta_{shared}}{(x_{i \in B})}})$} \hspace{10mm} $\triangleright$ evaluate supervised cnn outputs for  inputs
				\State{$z'_{i \in B} \gets f_{\theta_{unsup}}({f_{\theta_{shared}}{(x_{i \in B})}})$} \hspace{8mm}$\triangleright$ evaluate unsupervised cnn outputs for  inputs
				\State{$l_{i \in B} \gets -\frac{1}{|B|} \sum_{i \in B \cap S}{log f_{softmax}{(z_{i})}[y_{i}]}$}\hspace{10mm}$\triangleright$ supervised loss component
				\State{$l'_{i \in B} \gets \frac{1}{C|B|} \sum_{i \in B}{||z_{i} - z'_{i}||^{2}}$} \hspace{9mm} $\triangleright$ unsupervised loss component
				\State{$loss \gets l_{i \in B} + w(t) \times l'_{i \in B}$} \hspace{14mm} $\triangleright$ total loss 
				\State{update $\theta_{shared}$,  $\theta_{sup}$, $\theta_{unsup}$ using, e.g., ADAM} \quad $\triangleright$ update network parameters
 			 \EndFor
		\EndFor
	\Return{$\theta_{shared}$,  $\theta_{sup}$, $\theta_{unsup}$}
	\end{algorithmic}
	\label{Arg1_learning}
\end{algorithm*}

\section{Methodology}
We introduce the proposed model in the context of fake news detection. Suppose the training data consist of total $N$ inputs, out of which $M$ are labeled. The inputs are tweets that contain sentences related to fake news. We employ word embeddings \cite{mikolov2013distributed} to represent  input $x_{i}$  ($i \in {1 ... N }$) as ``images", where each row in the ``image" represents one word in the tweet as embeddings and the number of rows is the number of words in the tweet. $S$ is the set of labeled inputs, $|S| = M$. For every $i \in S$, we have a known correct label $y_{i} \in {1... C}$, where $C$ is the number of different classes. 

The framework of the proposed model and corresponding learning procedures are shown in Figure \ref{Fig1_framework} and Algorithm \ref{Arg1_learning}, respectively.  As shown in Figure \ref{Fig1_framework}, we evaluate the network for each training input $x_{i}$ with the supervised path and the unsupervised path, resulting in prediction vectors $z_{i}$ and $z'_{i}$, respectively. Then we utilize those two vectors to calculate the loss given by 

\begin{dmath}
Loss = -\frac{1}{|B|} \sum_{i \in B \cap S}{log f_{softmax}{(z_{i})}[y_{i}]} + w(t) \times \frac{1}{C|B|} \sum_{i \in B}{||z_{i} - z'_{i}||^{2}} \; ,
\label{Equ1_loss}
\end{dmath}

where $B$ is the minibatch in the learning process. The loss consists of two components. As illustrated in Algorithm \ref{Arg1_learning}, $l_{i}$ is the standard cross-entropy loss to evaluate the loss for labeled inputs only. On the other hand, $l'_{i}$, evaluated for all inputs, penalizes different predictions for the same training input $x_{i}$ by taking the mean squared error between $z_{i}$ and $z'_{i}$. To combine the supervised loss $l_{i}$ and unsupervised loss $l'_{i}$, we scale the latter by time-dependent weighting function $w(t)$ \cite{laine2016temporal} that ramps up, starting from zero, along a Gaussian curve. 

Although there are a few related works in the literature such as the $\Pi$ model \cite{laine2016temporal}, there exist significant differences. In the $\Pi$ model, it combined image data augmentation with dropout to generate two outputs, but it cannot be used to process language in NLP tasks. Furthermore, instead of using one path CNN, we apply two independent CNN to generate those two outputs.

\section{Experiment}


\subsection{Datasets}

PHEME dataset \cite{zubiaga2016analysing} is related to nine events whereas this paper only focuses on the five main events, namely, Germanwings-crash (GC), Charlie Hebdo (CH), Sydney siege (SS), Ferguson (FE), and Ottawa shooting (OS). It has different levels of annotations such as thread level and tweet level. We only adopt the annotations on the thread level and thus classify the tweets as fake or true.  The detailed distribution of tweets and classes is shown in table \ref{tab1_PHAME}.

\begin{table}
        \begin{center}
                \begin{tabular}{|l|rrr|}
                    \hline \textbf{Events} & \textbf{Tweets} & \textbf{Fake} & \textbf{True}\\ \hline
                    	    GC		& 4,651 	& 2,637	& 2,014 \\
                            CH		& 40,178  	& 7,697	& 32,481 \\
                            SS		& 25,221 	& 9,046	& 16,175 \\
                            FE		& 25,054	& 6,686	& 18,368 \\ 
                            OS		& 12,656	& 6,624	& 6,032 \\ 
                            \hline
                            \textbf{Total} 	& \textbf{107,760}	& \textbf{32,690}	& \textbf{75,070} \\  
                     \hline
                \end{tabular}
       \end{center}
        \caption{\label{tab1_PHAME} Number of tweets and class distribution in the PHEME dataset.}
\end{table}

\subsection{Experiment Setup}

The key hyperparameters for the proposed model are: 
Dropout: 0.5,
Minibatch size: 25,
Number of epochs: 200,
Optimizer: Adam optimizer,
and Maximum learning rate: 0.001.
In addition, we construct baselines with traditional machine learning (Naive Bayes, Decision Tree, Adaboost, Support Vector Machine (SVM)) and one deep learning model (bidirectional recurrent neural networks (BRNN) with LSTM), where we utilize \textit{tf-idf} to extract features from tweets for building traditional machine learning models and word embeddings to build BRNN. For BRNN, the key hyperparameters are: 
Dropout: 0.8,
Number of hidden layers: 2,
Number of neurons in hidden layers: 100,
Minibatch size: 64,
Number of epochs: 100,
Optimizer: Adam optimizer,
and Learning rate: 0.001.

\subsection{Evaluation}
We perform leave-one-event-out (LOEO) cross-validation \cite{kochkina2018all}, which is closer to the realistic scenario where we have to verify unseen truth and evaluate models using macro-averaged F-score as fake news detection on the PHEME dataset suffer from a class imbalance. We also employ Precision, Recall, and Fscore to exam the detailed performance on different events.

\subsection{Results}

Table \ref{tab_compare} compares the performance of the proposed model with those of baselines. It is observed both traditional machine learning models and BRNN  perform worse than the proposed model. There are several reasons. Firstly, the samples built by  \textit{tf-idf} are too sparse to learn for traditional models, even if we employ singular value decomposition (SVD) to mitigate the sparse problem. Secondly, the sample distribution between the two classes are not balanced for some events, which reduces the macro-average values of BRNN. Thirdly, the training data and testing datasets may not have similar distribution since we apply leave-one-event-out (LOEO) cross-validation to evaluate the model performance.

It is also observed that the proposed model (DTSL) could obtain promising performance with very few labeled data. Furthermore, when we increase the ratio of labeled data from 5\% to 10\%, the performance (Macro-Fscore) is improved as well. However, when we increase the ratio to 30\%, the performance drops about 3\%, which might be caused by different data distributions between training data and testing data.

We also present the detailed performance per event in Table \ref{tab_detail}. As shown in Table  \ref{tab_detail}, for the balanced data from events Germanwings-crash (GC) and Ottawa shooting (OS), their recalls are improved when increasing the ratio of labeled data. For the unbalance data from events Charlie Hebdo (CH) and Ferguson (FE), their Fscores are improved when increasing the ratio of labeled data. Specifically, the Fscore on the event Ferguson (FE) is worse than those of other events because this event content is different from those of other events significantly \cite{zubiaga2016learning}. 

\begin{table}[t!]
        \begin{center}
                \begin{tabular}{|l|rrr|}
                    \hline \textbf{Model} & \textbf{MP} & \textbf{MR} & \textbf{MF}\\ \hline
                    	   Naive Bayes			&  39.28\% 	&  45.16\%	&  41.24\% \\
                            Decision Tree			&  37.13\%	&  31.92\%	&  33.03\%  \\
                            AdaBoost			&  45.99\%	&  14.12\%	&  20.43\% \\
                            SVM 				&  55.53\%	&  7.39\%		& 12.56\% \\ 
                            \hline
                            BRNN 			&  44.61\%	& 36.86\%		& 35.85\% \\  
                            \hline
                            \textbf{DTSL  (5\%)}	&  59.73\%	& 58.69\%  	&  53.90\% \\
                            \textbf{DTSL  (10\%)}	&  56.05\%	& 79.44\%  	&  61.53\% \\
                             \textbf{DTSL 	(30\%)}	&  44.20\%	& 77.58\%		&  57.98\%\\
                    \hline
                \end{tabular}
        \end{center}
        \caption{\label{font-table} Comparing performances between the proposed model (DTSL) and baselines with three evaluation methods, namely, Macro-Precision  (MP), Macro-Recall (MR), and Macro-Fscore (MF). Specifically, we show the performance of DTSL with different ratios (\%) of labeled data.
        }
        \label{tab_compare}
\end{table}

\begin{table}
        \begin{center}
                \begin{tabular}{|l|rrr|}
                     \hline &\multicolumn{2}{c}{5\% Labeled Data}&\\
                    \hline \textbf{Events} & \textbf{Precision} & \textbf{Recall} & \textbf{Fscore}\\ \hline
                    	   GC		& 83.33\%		& 33.33\%		& 47.62\%  \\
                            CH		& 75.00\%  	& 60.00\%		&  66.67\%\\
                            SS		&  53.33\%	& 61.54\%		&  57.14\%\\
                            FE		& 13.64\%		& 60.00\%		& 22.22\%\\ 
                            OS		& 73.33\%		& 78.57\%		& 75.86\%\\  
                    \hline &\multicolumn{2}{c}{10\% Labeled Data}&\\
                    \hline \textbf{Events} & \textbf{Precision} & \textbf{Recall} & \textbf{Fscore}\\ \hline
                    	   GC		& 56.00\%		& 100.00\%	& 71.79\%  \\
                         CH		& 55.56\%  	& 50.00\%		&  52.63\%\\
                         SS		&  70.00\%	& 53.85\%		&  60.87\%\\
                         FE		& 25.00\%		& 100.00\%	& 40.00\%\\ 
                         OS		& 73.68\%		& 93.33\%		& 82.35\%\\  
                     \hline
                     &\multicolumn{2}{c}{30\% Labeled Data}&\\ 
                    \hline \textbf{Events} & \textbf{Precision} & \textbf{Recall} & \textbf{Fscore}\\ \hline
                    	   GC		& 44.00\%		& 100.00\%	& 61.11\%  \\
                            CH		& 55.56\% 	& 33.33\%		& 71.43\% \\
                            SS		& 50.00\% 	& 54.55\%		& 52.17\% \\
                            FE		& 33.33\%		& 100.00\%	& 50.00\% \\ 
                            OS		& 44.20\%		& 100.00\%	& 55.17\% \\ 
                     \hline
                \end{tabular}
       \end{center}
        \caption{\label{tab1_PRF} Detailed evaluation results generated with different portions of labeled data on five events.}
        \label{tab_detail}
\end{table}



\section{Conclusion and Future Work}
In this paper, a novel deep learning model is proposed for fake news detection in social media. 
Because of the fast propagation of fake news, timely detection is critical to mitigate their effects. However, usually very few data samples can be labeled in a short  time, which in turn makes the supervised learning models infeasible. Hence, 
a deep semi-supervised learning model is proposed. The two paths in the proposed model generate supervised loss (cross-entropy) and unsupervised loss (Mean Squared Error), respectively. Then training is performed by jointly optimizing these two losses. 
Experimental results indicate that the proposed model could detect fake news from PHEME datasets effectively by using limited labeled data and lots of unlabeled data. In the future, we plan to examine the proposed model on other NLP tasks such as sentiment analysis. 



\bibliography{acl2019}
\bibliographystyle{acl_natbib}

\end{document}